\documentclass{article}
\usepackage{natbib}
\usepackage{amsmath,amssymb,amsfonts,graphicx,dsfont}
\usepackage{graphicx}
\usepackage{amsfonts}
\usepackage{dsfont}
\usepackage{amssymb}
\usepackage{amsmath}
\usepackage{epsf}
\usepackage{graphicx,psfrag,color,pstcol,pst-grad}
\usepackage{latexsym}
\usepackage{calc}
\usepackage{multicol}
\usepackage{subfigure}
\usepackage{pst-tree}
\usepackage{dsfont}
\usepackage{rotate}
\usepackage{epsf}
\usepackage{dsfont}
\usepackage{amsfonts}
\usepackage{dsfont}
\usepackage{amssymb}
\usepackage{amsmath}
\usepackage{color}
\usepackage{multirow}
\usepackage{epsfig}
\usepackage{makeidx}

\def\OMIT#1{}

\def\RR{\mathbb{R}}

\def\NN{\mathbb{N}}

\newcommand{\BlackBox}{\rule{1.5ex}{1.5ex}}  

\newtheorem{theorem}{Theorem}
\newtheorem{lemma}[theorem]{Lemma}
\newtheorem{proposition}[theorem]{Proposition}

\newtheorem{corollary}[theorem]{Corollary}
\newtheorem{definition}[theorem]{Definition}

\def\undemi{\frac{1}{2}}

\def\Xcal{\mathcal{X}}

\def\Mcal{M_+^b(\Xcal)}
\def\Mpcal{M_+^1(\Xcal)}

\def\Ncal{\mathcal{N}}

\def\Lcal{\mathcal{L}}

\def\PSD{{\mathbf{P}_n}}
\def\PD{{\mathbf{P}_n^+}}
\def\Kappa{\mathcal{K}}

\def\RR{\mathbb{R}}

\def\NN{\mathbb{N}}

\def\Mol{\mol_+^1(\Xcal)}

\def\OMIT#1{}

 \DeclareMathOperator{\diag}{diag}

\DeclareMathOperator{\defi}{def}
\DeclareMathOperator{\tr}{tr}\DeclareMathOperator{\mol}{Mol}
 
\DeclareMathOperator{\defeq}{\overset{\defi}{=}}

\DeclareMathOperator{\mspec}{mspec}

\title{Kernels for Measures Defined on the Gram Matrix of their Support}
\author{Marco Cuturi\\Princeton University\\
\texttt{mcuturi@princeton.edu}}

\begin{document}
\maketitle

\begin{abstract}
We present in this work a new family of kernels to compare positive
measures on arbitrary spaces $\Xcal$ endowed with a positive kernel
$\kappa$, which translates naturally into kernels between histograms
or clouds of points. We first cover the case where $\Xcal$ is
Euclidian, and focus on kernels which take into account the variance
matrix of the mixture of two measures to compute their similarity.
The kernels we define are semigroup kernels in the sense that they
only use the sum of two measures to compare them, and spectral in
the sense that they only use the eigenspectrum of the variance
matrix of this mixture. We show that such a family of kernels has
close bonds with the laplace transforms of nonnegative-valued
functions defined on the cone of positive semidefinite matrices, and
we present some closed formulas that can be derived as special cases
of such integral expressions. By focusing further on functions which
are invariant to the addition of a null eigenvalue to the spectrum
of the variance matrix, we can define kernels between atomic
measures on arbitrary spaces $\Xcal$ endowed with a kernel $\kappa$
by using directly the eigenvalues of the centered Gram matrix of the
joined support of the compared measures. We provide explicit
formulas suited for applications and present preliminary experiments
to illustrate the interest of the approach.
\end{abstract}

\section{Introduction}
Defining meaningful kernels on positive measures is an important
issue in the field of kernel methods, as it encompasses the topic of
comparing histograms, bags-of-components or clouds of points, which
all arise very frequently in applications dealing with structured
data.

In the pioneering applications of support vector machines to
structured data, histograms were often treated as simple vectors and
used as such through the standard Gaussian or polynomial
kernels~\cite{joachims:2002a}. Yet, more adequate kernels which
exploit the specificities of histograms have been proposed since.
Namely, the fact that histograms are vectors with nonnegative
coordinates~\cite{hein05hilbertian}, and whose sum may be normalized
to one, that is cast as discrete probability measures and treated
under the light of information
geometry~\cite{Lafferty2005,leb06metric}. Since such histograms are
usually defined on bins which are not equally dissimilar, as is for
instance the case with color, words or amino-acid histograms,
further kernels which may take into account an a priori inter-bin
similarity where subsequently
proposed~\cite{KonJeb03,cuturi05semigroup,hein05hilbertian} as an
attempt to include with more accuracy a prior knowledge on the
considered components, through the knowledge of a prior kernel
$\kappa$ for instance.

In this paper we investigate further such kind of kernels between
two measures, which can conveniently describe the similarity between
two clouds of points by only considering their Gram matrices. In
this sense we reformulate and extend the results
of~\cite{cuturi05semigroup} whose framework we briefly recall:

The set $\Mcal$ of bounded positive measures on a set $\Xcal$ is a
cone, and from a more elementary algebraic viewpoint a
semigroup\footnote{In this paper, a semigroup will be a non-empty
set $S$ endowed with a commutative addition, such that for $s,t\in
S$, their sum $s+t=t+s\in S$, and a neutral element $e$ such that
$s+e=s$}. In that sense, a natural way to define kernels suited to
the geometry of $\Mcal$ is to study the family of semigroup
functions on $\Mcal$, as introduced in~\cite{berg84harmonic}, that
is real-valued functions $\psi$ defined on $\Mcal$ such that the map
$(\mu,\mu')\mapsto \psi(\mu+\mu')$ is either positive or negative
definite. The Jensen-divergence, which is computed through the
entropy of the mixture of two measures is such an example, as
recalled in~\cite{hein05hilbertian}.

Given the complexity of evaluating entropies for finite samples, it
is shown in~\cite{cuturi05semigroup} that similar quantities can be
defined for measures by only taking into account the variance
matrix~$\Sigma(\mu+\mu')$~of the mixture of two measures. This has
two clear advantages. First, variances are easy to compute given
atomic measures, that is measures with finite support, which are
usually considered in most applications. Second, the eigenspectrum
of the variance matrix of an (atomic) probability measure is known
to be the same as, up to zero eigenvalues and an adequate
centralization, the eigenspectrum of the dot-product matrix of the
support of the same measure. This fact paves the way to consider
kernels defined on Gram matrices rather than on variance matrices,
regardless of the structure of $\Xcal$, as was first hinted
in~\cite{KonJeb03}.

More precisely, the authors of ~\cite{cuturi05semigroup} first prove
that for a variance matrix $\Sigma(\mu+\mu')$, the determinant
$|\frac{1}{\eta}\Sigma(\mu+\mu')+I_n|^{-\undemi}$ for $\eta>0$ is a
positive definite (p.d.) kernel between two measures $\mu,\mu'$ on
an Euclidian space $\Xcal$ of dimension $n$. Second, they prove that
this quantity can be cast into a reproducing kernel Hilbert space
(rkHs) associated with a kernel $\kappa$ on $\Xcal$, regardless of
the nature of $\Xcal$, by using directly a centered Gram matrix
$\Kappa_{\mu,\mu'}$ of all elements contained in the support of both
$\mu$ and $\mu'$.

We are interested in this paper in characterizing other functions
$\varphi$ defined on matrices such that
\textit{(i)}~$\mu,\mu'\mapsto \varphi(\Sigma(\mu+\mu'))$ is either
positive or negative definite, \textit{(ii)}~$\varphi$ is
spectral\footnote{A function $f$ defined on symmetric matrices is
spectral, or orthogonally invariant, if for any real $n\times n$
orthogonal matrix $H$, that is such that $HH^\top=I_n$,
$f(HAH^\top)=f(A)$. In that case $f$ only depends on the
eigenspectrum of $A$. See~\cite{borwein1}} and
\textit{(iii)}~$\varphi$ is invariant to the addition of null
eigenvalues, that is, for two square p.d. matrices $A,B$ which may
not have the same size, $\varphi(A)=\varphi(B)$ if $A$ and $B$ have
the same \emph{positive} eigenvalues taken with their multiplicity,
regardless of the multiplicity of $0$ in their eigenspectrum. It is
easy to check that both $|\frac{1}{\eta}\cdot+I|$ and the trace
fulfill condition \textit{(iii)}.

If $\varphi$ satisfies condition \textit{(i)} and \textit{(ii)}, we
call the composed function $\psi=\varphi\circ\Sigma$ a
\emph{semigroup spectral positive (resp. negative) definite}
(s.s.p.d., resp. s.s.n.d.) function on $\Mcal$. Note that the task
of defining such functions $\psi$ is \emph{not} equivalent to
defining directly positive or negative definite functions $\varphi$
on the semigroup of p.d. matrices, since the underlying semigroup
operation is the addition of measures and not that of the variance
matrices of the measures, as recalled in
Equation~\eqref{eq:sumvariance}. When $\varphi$ is further invariant
to null eigenvalues \textit{(iii)} $\psi$ can be cast in Hilbert
spaces of infinite dimensions to compare degenerated variance
operators, which will be in the context of this paper an rkHs built
on $\Xcal$ through a kernel $\kappa$.

This paper is structured as follows: we introduce in
Section~\ref{sec:ssf} an alleviated formalism for semigroup
functions, and propose a general link between s.s.p.d. functions and
the Laplace transform of functions defined on matrices in
Section~\ref{sec:zonal}. We review then in Section~\ref{sec:low}
different s.s.p.d. functions, notably a function which satisfies
criteria \textit{(iii)} and which does not requite any
regularization. We provide explicit formulas and test the kernel
derived from such a function on a benchmark classification task
involving handwritten digits in Section~\ref{sec:experiments}.

\section{Semigroup Functions on Bounded Subsets of $\Mcal$}\label{sec:ssf}
We consider $\Xcal$, an Euclidian space of dimension $n$ endowed
with Lebesgue's measure and restrict $\Mcal$ to measures with finite
first and second moments. In such a case, the variance of a measure
$\mu$ of $\Mcal$ can be defined as:
$$
\Sigma(\mu)= \mu[xx^\top]-\mu[x]\mu[x]^\top.
$$
Writing $\bar\mu$ for $\mu[x]$, we recall an elementary result for
two measures $\mu,\mu'$ of $\Mcal$,
\begin{equation}\label{eq:sumvariance}
\Sigma(\mu+\mu')=\Sigma(\mu)+\Sigma(\mu')-\left(\bar\mu\bar\mu'^\top+\bar\mu'\bar\mu^\top\right),
\end{equation}
which highlights the nonlinearity of the variance mapping.

We write $\PSD$ for the cone of real, symmetric and positive
semidefinite matrices, and $\PD$ for its subset of (strictly) p.d.
matrices. In this paper, the assumption that for a measure $\mu$ its
variance $\Sigma(\mu)$ is in $\PSD$ is crucial for most
calculations, and this is ensured for sub-probability measures, that
is is measures $\mu$ such that $|\mu|=\mu(\Xcal)\leq1$, since we
then have that
\begin{equation}\label{eq:okpourprobas}
\Sigma(\mu)=\mu[\left(x-\bar\mu\right)\left(x-\bar\mu\right)^\top]+\left(1-|\mu|\right)\bar\mu\bar\mu^\top\in\PSD.
\end{equation}
Furthermore, we will also need the identity
$\Sigma(\mu)=\mu[\left(x-\bar\mu\right)\left(x-\bar\mu\right)^\top]$
in order to make the link between the dot-product matrix of the
support of $\mu$ and its variance matrix, which is why we restrict
our study to probability measures $\Mpcal$. $\Mpcal$ is not,
however, a semigroup, since it is not closed under addition, due to
the constraint on $|\mu|$. To cope with this contradiction, that is
to use semigroup-like functions of the type
$(\mu,\mu')\rightarrow\psi(\mu+\mu')$ where $\psi$ is only defined
on a subset of the original semigroup, and where this subset may not
be itself a semigroup, we define the following extension to the
original definition of semigroup functions which, although
technical, is also useful to recall the actual definitions of
positive and negative definiteness for semigroup functions.
\begin{definition}[Semigroup kernels on subsets]\label{def:extensions}
Let $(S,+)$ be a semigroup  and $U\subset S$ a nonempty subset of
$S$. A function $\psi:U\rightarrow \RR$ is a p.d.~(resp. n.d.)
semigroup function on $U$ if
$$
\sum_{i,j}c_i c_j \psi(s_i+s_j) \geq 0 \;\;(\text{resp} \leq 0)
$$
holds for any $n \in \NN$; any $s_1,\ldots,s_n \in S$ such that
$s_i+s_j\in U$ for $ 1\leq i\leq j \leq n$; and any $c_1 \ldots,c_n
\in \RR$ (resp. with the additional condition that $\sum_i c_i=0$)
\end{definition}
In practice, stating that a function $\psi$ defined on the subset
$\Mpcal$ is positive (resp. negative) definite is equivalent to
stating that the kernel for two elements $\mu,\mu'$ of $\Mpcal$
defined as
$$(\mu,\mu') \mapsto
\psi\left(\frac{\mu+\mu'}{2}\right)$$ is positive (resp. negative)
definite. Finally, we write $\Sigma^{-1}(\mu)$ for
$(\Sigma(\mu))^{-1}$ when appropriate.

\section{Laplace Transforms of Matrix Functions and s.s.p.d. functions}\label{sec:zonal}

We show in this section how s.s.p.d. functions on $\Mpcal$ can be
defined through the Laplace transform of a nonnegative-valued
function defined on the cone $\PD$, through the following lemma.
\begin{lemma}\label{lem:tracepd}
For any $S\in\PSD$, the real-valued function defined on $\Mpcal$,
$$\mu\mapsto \langle\Sigma(\mu),S\rangle$$ is
a negative definite semigroup function.
\end{lemma}
\textit{Proof.} For any $k\in\NN$, any $c_1,\ldots,c_k\in\RR$ such
that $\sum_i c_i=0$ and any $\mu_1,\ldots,\mu_k\in\Mpcal$ such that
$\mu_i+\mu_j\in\Mpcal$, we have using
Equation~\eqref{eq:sumvariance} that
\begin{multline*}
\sum_{i,j} c_i c_j
\left\langle\Sigma\left(\mu_i+\mu_j\right),\,S\,\right\rangle =
\Biggl\langle \sum_{i,j} c_i c_j \Big(
\Sigma(\mu_i)+\Sigma(\mu_j)-\left(\bar\mu_i\bar\mu_j^\top+\bar\mu_j\bar\mu_i^\top\right)\Big)\,,\,S\Biggl\rangle\\
= - \left\langle \sum_{i,j} c_i c_j
\left(\bar\mu_i\bar\mu_j^\top+\bar\mu_j\bar\mu_i^\top\right)\,,\,S\right\rangle=
-{2}\sum_{i,j} c_i c_j \bar\mu_i^\top S \bar\mu_j \leq 0.\BlackBox
\end{multline*}

Note that this function is actually n.d. for all measures of
$\Mcal$, regardless of their total weight $|\mu|$. The case $S=I_n$
yields the simple function
$\psi_{\tr}\defeq\mu\mapsto\tr\Sigma(\mu)$, which provides
interesting results in practice, and boils down to a fast kernel on
clouds of points, which we will review briefly in
Section~\ref{sec:experiments}. For any nonnegative-valued function
$f:\PD\rightarrow \RR^+$ defined on the cone of p.d. matrices, we
write
\begin{equation}\label{eq:laplacetrans}
\Lcal f(Z)=\int_{S\in\PD}e^{-<Z,S>}f(S)dS
\end{equation}
for the Laplace transform of $f$ evaluated in $Z\in\PD$, when the
integral exists.

\begin{proposition}\label{prop:posdef}
For any spectral function $f:\PD\rightarrow \RR^+$, the mapping
$$
\mu\mapsto \Lcal f\left(\Sigma(\mu)\right)
$$
defined for all measures $\mu\in\Mpcal$ such that
$\Sigma(\mu)\in\PD$ is a s.s.p.d.~function.
\end{proposition}
\textit{Proof.} The integral when it exists is a sum of
p.d.~semigroup functions through Schoenberg's theorem~\citep[Theorem
3.2.2]{berg84harmonic}, and is hence p.d.\BlackBox

Laplace transforms of functions defined on matrices is an extensive
subject and we refer to~\cite[Section 4]{mathai} for a short survey.
In the case where $f=1$ we recover the characteristic function of
the cone $\PD$, and its logarithm, $\ln\Lcal1(A)= C -
\frac{n+1}{2}\log|A|$, is known as the universal
barrier~\cite{guler4} of the cone $\PD$, with numerous applications
in convex optimization.

We recall now a well-known result of multivariate analysis based on
zonal polynomials (see ~\cite{takemura,mathaiprovost} for an
exhaustive presentation of these), which may not, however, be of
immediate use for an application in kernel methods. To be short,
zonal polynomials $C_\alpha(A)$ are polynomials in the eigenvalues
of a matrix $A$ with positive coefficients~\cite[Remark
4.3.6]{mathaiprovost}, and thus nonnegative-valued spectral
functions, indexed by the partitions $\alpha$ of an integer $a$.
Namely, for $a\in\NN$, we write $\alpha=(a_1,a_2,\ldots,a_n)$ for a
partition of $a$ into not more than $n$ parts, that is
$a_1+a_2+\cdots+a_n=a$ and $a_1\geq a_2\geq \cdots \geq a_n$. The
following result follows from~\cite[Theorem 4.4.1]{mathaiprovost}
where we have dropped constants which only depend of $n$ and
$\alpha$ for more readability:
\begin{corollary}\label{prop:zonalpoly}
Given $a\in\NN$ and a partition $\alpha$ of $a$, the real-valued
zonal kernel $\psi_\alpha$ is a s.s.p.d.~function on $\Mpcal$, with
$$
\psi_\alpha:\mu\mapsto\frac{C_\alpha(\Sigma^{-1}(\mu))}{|\Sigma(\mu)|^{-\frac{1}{2}n}},
$$
through the identity
$\int_{S\in\PD}e^{-<\Sigma,S>}|S|^{t-\undemi(n+1)}C_\alpha(S)
dS\propto|\Sigma|^{-t}C_\alpha(\Sigma^{-1})$, for $t>\undemi(n-1)$.
\end{corollary}
Actual expressions for zonal polynomials of order $a\leq 10$ are
currently known, and the use of Wishart densities for $f$ can be
seen as a special case of such evaluations. It is also clear that
finite and infinite linear combinations of such zonal kernels, with
the speculation that they might be a useful basis for a subcategory
of s.s.p.d. functions, can be carried out in the spirit of equations
provided in~\cite[Lemmas 4.4.5\&6]{mathaiprovost} and yield
convenient formulas, such as
$$
\sum_{a=k}^{\infty}\sum_{\alpha}\psi_{\alpha}:\mu\mapsto
\frac{e^{\tr\Sigma^{-1}(\mu)}(\tr\Sigma^{-1}(\mu))^k}{|\Sigma(\mu)|^{-\undemi
n}},
$$
which is a s.s.p.d. function for any $k\geq0$. However, the weak
point of these expressions when used in our setting is that they
tend to be extremely degenerated when the eigenspectrum of $\Sigma$
vanishes, due to the high power of the denominator and to the fact
that the eigenspectrum of $\Sigma^{-1}$, not $\Sigma$, is considered
implicitly. Hence, we do not see at the moment how one would obtain
expressions satisfying condition \textit{(iii)}, even through the
use of regularization. To handle this problem, we focus in the next
section on degenerated integrations, that is we consider an
extension of the Laplace transform setting defined in
Equation~\eqref{eq:laplacetrans} to degenerated functions $f$
defined on families of semidefinite matrices of $\PSD$.

\section{Degenerated Integrations on Semidefinite Matrices of Rank $1$}\label{sec:low}
We restrict the integration domain to only consider the subspace of
$\PSD$ of matrices of rank $1$, that is matrices of the form
$yy^\top$ where $y\in\RR^n$. The Euclidian norm $y^\top y$ of $y$ is
the only positive eigenvalue of $yy^\top$ when $y\neq 0$, hence only
real-valued functions of $y^\top y$ can be spectral. Following the
proof of Proposition~\ref{prop:posdef}, and for any
nonnegative-valued function $g:\RR^+\rightarrow\RR^+$, we observe
thus that
\begin{equation}\label{eq:int_elem}
\psi_g:\mu\mapsto\int_{\RR^n}e^{-y^t\Sigma(\mu)y}g(y^\top y)dy
\end{equation}
is a s.s.p.d.~function on $\Mpcal$, noting simply that
$\tr(\Sigma(\mu) y y^\top)= y^t\Sigma(\mu) y$. We start our analysis
with a simple example for $g$, which can be computed in close form.

\subsection{The case $g:x\mapsto x^i$} For a matrix $A\in\PD$ such
that $\mspec A=\{\lambda_1,\ldots,\lambda_n\}$, we set
$\gamma_0(A)=1$ and write for $1\leq i \leq n$,
$$\gamma_i(A)\defeq\sum_{|j|=i}\frac{\prod_{k=1}^n\Gamma(j_k+\frac{1}{2})}{\lambda_1^{j_1}\cdots\lambda_n^{j_n}}$$ where the
summation is taken over all families $j\in\NN^n$ such that the sum
of their elements $|j|$ is equal to $i$. Writing $\sigma_n$ for
$(2\pi^{-\undemi})^\frac{n}{2}$ we have with these notations that
for all $1\leq i \leq n$,
\begin{corollary}\label{cor:yytop} The function
$\psi_i:\mu\mapsto \sigma_n
\sqrt{\gamma_n}\cdot\gamma_i\left(\Sigma(\mu)\right)$ is a
s.s.p.d.~functions on $\Mpcal$.
\end{corollary}
\textit{Proof.} Let $\mu\in\Mpcal$, and write
$\mspec\Sigma(\mu)=\{\lambda_1,\ldots,\lambda_n\}$. Then by an
appropriate base change we have for $g_i:x\mapsto x^{i}$, $i\leq n$,
$$
\begin{aligned}
\psi_{g_i}&(\mu) \;=\; \int_{\RR^n}e^{-\sum_{k=1}^{n}\lambda_k y_k^2} \,\left(\sum_{k=1}^{n} y_k^2\right)^i dy =\;\int_{\RR^n}e^{-\sum_{k=1}^n\lambda_k y_k^2}\sum_{|j|=i} \prod_{k=1}^n y_k^{2j_k} dy\\
&=\; \sum_{|j|=i} \prod_{k=1}^{n} \int_{\RR}e^{-\lambda_k y_k^2}
y_k^{2j_k}dy_k \;=\; \sum_{|j|=i} \prod_{k=1}^{n}
\Gamma(j_k+\frac{1}{2})
\lambda_{k}^{-j_k-\frac{1}{2}}\;=\sigma_n\sqrt{\gamma_n}\cdot\gamma_i\left(\Sigma\left(\mu\right)\right).\BlackBox
\end{aligned}
$$
The inverse generalized variance is recovered as $\psi_0$. We refer
now to Lancaster's formulas~\citep[p.320]{Bernstein} to express more
explicitly the cases $i=1,2,3$, where we write $\Sigma$ for
$\Sigma(\mu)$:
$$\begin{aligned}
\psi_1(\mu)=&\frac{\sigma_n}{\sqrt{|\Sigma|}}\left[\tr\Sigma^{-1}\right],\,\,\,\psi_2(\mu)= \frac{\sigma_n}{\sqrt{|\Sigma|}}\left[(\tr\Sigma^{-1})^2+2\tr\Sigma^{-2}\right],\\
\psi_3(\mu)=&\frac{\sigma_n}{\sqrt{|\Sigma|}}\left[(\tr\Sigma^{-1})^3+6(\tr\Sigma^{-1})(\tr\Sigma^{-2})+8\tr\Sigma^{-3}\right].
\end{aligned}$$
Although the functions $\psi_i$ are s.s.p.d., they are mainly
defined by the lowest eigenvalues of $\Sigma(\mu)$. These functions
can all be regularized, by adding a weighted identity matrix $I_n$
to $\Sigma$, while still preserving their positive definiteness as
can be easily justified by using the functions $g_i(x)=e^{-x} x^i$
to penalize for large values of $y^\top y$. In such a case however,
and to the notable exception of $\psi_0$, this regularization
prevents the above functions to be invariant to the addition of a
zero eigenvalue to the spectrum of $\Sigma(\mu)$. Intuitively, this
degeneracy is due to the fact that we integrate on the whole or
$R^n$, notably on $\ker\Sigma(\mu)$, where the contribution of
$\exp(-y^\top\Sigma(\mu)y)$ is infinite. We propose to solve this
issue by considering more specifically the contribution of each
sphere $\{y|y^\top y = t\}$ to the overall summation in the case
where $g=1$.

\subsection{The case $g:x\mapsto \delta_{t}$ and its variants}\label{sec:ruben} The question of integrating
$\exp(-y^\top\Sigma y)$ over compact balls $\{y\in\RR^n \,|\, y^\top
y\leq t\}$ or spheres $\{y\in\RR^n \,|\, y^\top y=t\}$ is closely
related to the evaluation of the distribution of quadratic forms in
normal variates~\cite{mathaiquadratic}. Given a matrix $Q\in\PSD$
and a random vector $y$ in $\RR^n$ following a normal law
$\Ncal(m,V)$ with $V\in\PD$, the density $h[Q,V,m]$ of the values of
$y^\top Qy$, that is

\begin{equation}
\label{eq:normal} h[Q,V,m](t)
dt = (2\pi)^{-\frac{n}{2}}|V|^{-\frac{1}{2}} \int_{t< y^\top Q y<
t+dt} e^{-(y-m)^\top V^{-1}(y-m)}\,dy,
\end{equation}

can be evaluated explicitly in terms of the eigenvalues of
$V^{\frac{1}{2}}QV^{\frac{1}{2}}$ through series
expansions~\cite{mathaiquadratic}, as well as the distribution of
$h$ and its Laplace transform. We note that these expressions are
similar to that of the elementary contribution of the sphere
$\{y^\top y=t\}$ when $g=1$,

\begin{equation}\label{eq:laplace}
f_\mu(t)dt=\int_{t<y^\top
y<t+dt}e^{-y^\top\Sigma(\mu)y}\,dy.\end{equation}

The difference between expressions~\eqref{eq:laplace}
and~\eqref{eq:normal} is that the variance $\Sigma^{-1}(\mu)$ may be
potentially infinite if Equation~\eqref{eq:laplace} is directly
translated in terms of Equation~\eqref{eq:normal}, while the
normalization term in $\sqrt{|\Sigma(\mu)|}$ does not appear in the
s.s.p.d. function of Equation~\eqref{eq:laplace}. It turns out that
these two problems can be easily canceled out. We consider the
Laplace transform of $f_\mu(t)$ defined as
$L_\mu(s)=\int_{t\geq0}e^{-st}f_\mu(t)dt$, which is trivially a
s.s.p.d. as a sum of elementary s.s.p.d. functions.

\begin{lemma}\label{lem:mathai}
For $\mspec\Sigma(\mu)=\{\lambda_1,\ldots,\lambda_n\}$, define the
sequences $d_k=\frac{1}{2}\sum_{j=1}^n \lambda_j^k$ for $k\geq 1$,
$c_0=1$ and $c_k=\frac{1}{k}\sum_{r=0}^{k-1}d_{k-r}c_r$ for $k\geq
1$. Then,
$$
f_\mu(t)=\pi^{\frac{n}{2}}\sum_{k=0}^{\infty}(-1)^k c_k\,
\frac{t^{\frac{n}{2}+k-1}}{\Gamma(\frac{n}{2}+k)};\,\quad
L_\mu(s)=\pi^{\frac{n}{2}}\sum_{k=0}^\infty (-1)^k c_k
\,s^{-\frac{n}{2}+k}.
$$
\end{lemma}
\textit{Proof.} Take the exact value computation for
$$\frac{\pi^{\frac{n}{2}}}{|\Sigma(\mu)+2\eta I_n|^{\frac{1}{2}}}
h[I_n,(\frac{1}{2}\Sigma(\mu)+\eta I_n)^{-1},0](t)=\int_{y^\top
y=t}e^{-y^\top(\Sigma(\mu)+\eta I_n)y}\,dy.
$$ and let $\eta\rightarrow 0$ to obtain through~\cite[Lemma 4.2b.2
\& Theorem 4.2b.1]{mathaiquadratic} the corresponding formulas, both
for $f_\mu$ and $L_\mu$\BlackBox

\begin{proposition}
By considering $s=1$ in $L_\mu(s)$ from Lemma~\ref{lem:mathai}, and
noting that $d_k=\frac{1}{2}\tr(\Sigma(\mu)^k)$ we have that
$$\mu\mapsto \psi_M(\mu)=\sum_{k=0}^\infty (-1)^k c_k$$ is a s.s.p.d.
function on $\Mpcal$, invariant to the addition of null eigenvalues,
defined when the spectrum of $\Sigma(\mu)$ is strictly upper-bounded
by $1$.
\end{proposition}

Although all terms $d_k$ are invariant by the addition of a null
eigenvalue, $f_\mu$ cannot fulfill condition $(iii)$ because of the
numerator in $\Gamma(\frac{n}{2}+k)$ which depends on both the
dimension $n$ and the summation variable $k$.

We refer to the proof~\cite[Theorem 7]{cuturi05semigroup} to show
that for an atomic measure $\mu$ such that $|\mu|=1$, that is
$\mu\in\Mol$, $\psi_M(\mu)$ can be either expressed in terms of the
spectrum of its variance matrix or in the spectrum of its centered
dot-product matrix. Thus, if $\Xcal$ is now an arbitrary space
endowed with a kernel $\kappa$, the centered Gram matrix
corresponding to the support of $\mu$ can be used directly as an
input for a s.s.p.d. function such as $\psi_M$ in order to define a
p.d. kernel on $\Mol$. We detail in Section~\ref{sec:experiments}
the formulation of such a kernel, along with two other examples.

Finally, as a closing remark for this section, note that the
representation proposed in Proposition~\ref{prop:posdef} is not
exhaustive to our knowledge, and should not be confused with the
integral representation of semigroup p.d.~functions as sums of
semicharacters, studied in~\cite{cuturi05semigroup}. First, the
functions $\mu\mapsto e^{-\langle\Sigma(\mu),S\rangle}$ are not
semicharacters\footnote{Semicharacters of a semigroup $S$ are,
following the definition of~\cite{berg84harmonic}, real-valued
functions $\rho$ on $S$ such that for $s,t\in S$,
$\rho(s+t)=\rho(s)\rho(t)$} of the semigroup $\Mpcal$, since
$$
e^{-\langle\Sigma(\mu+\mu'),S\rangle}\neq e^{-\langle\Sigma(\mu)
+\Sigma(\mu'),S\rangle }
$$
in the general case. Second, the class of functions considered
through Lemma~\ref{lem:tracepd} is far from characterizing all
semigroup n.d. functions on $\Mpcal$ since, through~\citep[Corrolary
3.2.10, p.78]{berg84harmonic}, we have that for any $S\in\PSD$ and
$0<\beta<1$ both
$$\mu\mapsto \langle\Sigma(\mu),S\rangle^\beta, \quad\text{and}\,\,\, \mu\mapsto \ln(1+\langle\Sigma(\mu),S\rangle),$$ are
semigroup n.d.~functions. Note that if we use for $y\in\RR^n$ and
$m\geq 1$ a n.d.~function of the type
$$
\mu\mapsto
\frac{m+n}{2}\ln\left(1+\frac{1}{m}y^\top\Sigma(\mu)y\right),$$ and
exponentiate it in the spirit of Equation~\eqref{eq:int_elem}, we
recover the integration of the Student multivariate distribution for
vectors of $\RR^n$, which boils down again to a kernel that is
proportional to $\psi_0$.

\section{Explicit Formulas for Atomic Measures and Experiments}\label{sec:experiments}
Given two clouds of weighted points $\gamma=(x_i,a_i)_{i=1}^d$ and
$\gamma'=(y_i,b_i)_{i=1}^{d'}$, we show how to compute three
different kernels which satisfy condition \textit{(iii)}, namely
$\psi_{\tr}$, $\psi_0$ and $\psi_M$, and compare them by studying
their performance on a multiclass classification task.

\subsection{Formulations for Clouds of Points}
The mixture of $\gamma$ and $\gamma'$ can be expressed as $\gamma''=
\{(x_i,\undemi a_i)_{i=1}^d,(y_{j},\undemi b_{j})_{j=1}^{d'}\}$. By
writing $d''=d+d'$ and
$$
\Kappa_\gamma=[\kappa(x_i,x_j)]_{i,j\leq d}, \,\,\,\,
\Kappa_\gamma=[\kappa(y_i,y_j)]_{i,j\leq d'}, \text{and}\,\,
\Kappa_*=[\kappa(x_i,y_j)]_{i\leq d,\,j\leq d'}\\
$$
we can further express the $d''\times d''$ Gram matrix of the
mixture $\gamma''$ as
$$
\Kappa_{\gamma''}=\left(%
\begin{array}{cc}
  \Kappa_{\gamma} & \Kappa_* \\
  \Kappa_*^\top & \Kappa_{\gamma'} \\
\end{array}%
\right).
$$
As can be seen in~\cite{cuturi05semigroup}, the spectrum of the Gram
matrices cannot be taken as such since they require a centralization
of the form
\begin{equation}\label{eq:central}
\tilde{\Kappa}_{\gamma''}=
(I_{d''}-\mathds{1}_{d'',d''}\Delta_{\gamma''}) \Kappa_{\gamma''}
(I_{d''}-\Delta_{\gamma''}\mathds{1}_{d'',d''} )\Delta_{\gamma''},
\end{equation}
with $\Delta_{\gamma''}=\diag(\undemi a_i,\undemi b_j)$ and
$\mathds{1}_{d'',d''}$ is the $d''\times d''$ matrix of ones. The
explicit formulas for the considered kernels, which we do not
normalize and propose with bandwidth parameters are thus
$$
\begin{aligned}
k_{\tr}(\gamma,\gamma')&=e^{-\frac{1}{t}\tr\tilde{\Kappa}_{\gamma''}}\\
k_0(\gamma,\gamma')&=|\frac{1}{\eta}\tilde{\Kappa}_{\gamma''}+I_{d''}|^{-\undemi}\\
k_M(\gamma,\gamma')&=\sum_{k=0}^\infty (-1)^k c_k\,,\,\,\,  \text{
with } c_0=1, c_k=\frac{1}{k}\sum_{r=0}^{k-1}d_{k-r}c_r \text{ and }
d_k=\frac{1}{2}\tr([\delta\tilde{\Kappa}_{\gamma''}]^k)
\end{aligned}
$$
with $t>0, \eta>0$ and $\delta>0$ such that
$\delta<1/\rho(\tilde{\Kappa}_{\gamma''})$, where for a matrix $A\in
\PSD$ such that $\mspec A=\{\lambda_1,\ldots,\lambda_n\}$, $\rho(A)$
is the spectral radius of $A$, that is $\max_{1\leq i\leq
n}\lambda_i$.

We discuss now possible values for $\delta$ which will ensure that
$\delta<1/\rho(\tilde{\Kappa}_{\gamma})$ for any cloud-of-points
$\gamma$ and any kernel $\kappa$ upper-bounded by one, that is
$sup_{x\in\Xcal}|\kappa(x,x)|\leq 1$. Through
Equation~\eqref{eq:central}, one can obtain that for any cloud of
points $\gamma=(x_i,a_i)_{i=1}^{d}$,
$$
\rho(\tilde{\Kappa}_\gamma)\leq [\max(d\cdot a_{\max}-1,1)]^2 d
\cdot a_{\max}
$$
where we write $a_{\max}$ for the maximal weight of $\gamma$ and we
have bounded $\rho(\Kappa_{\gamma})$ by $d$, which corresponds to
the case $\Kappa_{\gamma}=\mathds{1}_{d,d}$. Thus, any factor
$\delta$ chosen so that
$$
\delta<\frac{1}{[\max(d\cdot \omega-1,1)]^2 d \cdot \omega}
$$
can be used to compare families of clouds of points whose maximal
weights do not exceed $\omega$ and maximal size does not exceed
$\undemi d$. In the case where these clouds are bounded between
$d_{\min}$ points (with weight $1/d_{\min}$) and $d_{\max}$ points,
this condition is ensured for
$\delta\leq(\frac{d_{\min}}{d_{\max}})^3$, which is far from being
optimal in practical cases since the values of $\kappa$ are more
likely to be better distributed in the [0,1] range. This shows
however that if we compare clouds of similar size $\delta$ can be
equal to 1, and possibly above depending on the kernel $\kappa$
which is used. We leave for future work the study of the convergence
of the series $\sum_{k=0}^N (-1)^k c_k$ corresponding to the
evaluation of $k_M$, although we note that in the practice of our
experiments very few iterations (that is $N$ set between 10 and 20)
are sufficient to converge to the limit value, which reduces
considerably the overall computation cost with respect to a
straightforward eigenvalue decomposition of
$\tilde{\Kappa}_{\gamma''}$. Indeed, as is the case with the inverse
generalized variance, this would have a cost of the order of $d^3$
while $N$ computations of the traces
$\tr[\delta\tilde{\Kappa}_{\gamma''}]^k$ only grow in complexity
$Nd^2$. It would be wise, however, to let $N$ depend adaptively on
the convergence of $\tr([\delta\tilde{\Kappa}_{\gamma''}]^k)$ to 0,
which is very much conditioned by the observed spectrum for
$\kappa$.

\subsection{Experiments on MNIST handwritten digits}
We use the Experimental setting of~\cite{KonJeb03}, also used
in~\cite{cuturi05semigroup} to compare the three previous kernels,
namely, we sample 1.000 images from the MNIST database, that is 100
images per digit, and sample randomly clouds-of-pixels to compare
such digits using the three kernels described above. The images,
which are actually $28\times28$ matrices, are considered as
clouds-of-pixels in the $[0,1]^2$ square, and we use a Gaussian
kernel of width $\sigma=0.1$ to evaluate the similarity between two
pixels through $\kappa$, and use a three fold cross validation with
five repeats to evaluate the performances of the kernels. The
preliminary results shown in Table~\ref{tab:results} show that the
kernel $\psi_M$ is competitive with both $\psi_{\tr}$ and the
inverse generalized variance, which was itself shown to be effective
with respect to other kernels in~\cite{cuturi05semigroup}, such as
simple polynomial and Gaussian kernels.

\begin{table}[htbp]
\begin{center}
\begin{tabular}{|c||c|c|c|c|}
\hline Sample Size &  $\psi_0,\eta=0.01$ & $\psi_{\tr},t=0.1$ & $\psi_M, \delta=1$ \\
 \hline
40 pixels & 16.2  &  28.6  &  20.62
 \\
50 " & 14.7  &  16.47   &15.84
\\
60 " & 14.5  &  14.97  & 13.52
\\
70 " & 13.1   & 11.3 &   13
\\
80 " & 12.8  &  10.8  &  12.4\\
\hline\end{tabular}\caption{Misclassification rate expressed in
percents for the 3 s.s.p.d. functions used on a benchmark test of
recognizing digits images, with 40 to 80 black points sampled from
the original images.}\label{tab:results}
\end{center}
\end{table}

\bibliographystyle{alpha}
\bibliography{/Users/mcuturi/Documents/REDACTION/COMMON/bib_short}
\end{document}